\documentclass[11pt,a4paper]{article}
\usepackage{times,latexsym}
\usepackage{url}
\usepackage[T1]{fontenc}

\usepackage[acceptedWithA]{tacl2021v1}
\usepackage{graphicx}
\usepackage{booktabs}
\definecolor{rowgray}{HTML}{EFEFEF}
\definecolor{darkgray}{HTML}{D1D1D1}
\usepackage{multirow}
\usepackage{amsmath}
\usepackage{nicematrix}
\usepackage{natbib}

\usepackage{xspace,mfirstuc,tabulary}

\newif\iftaclinstructions
\taclinstructionsfalse % AUTHORS: do NOT set this to true
\iftaclinstructions
\renewcommand{\confidential}{}
\renewcommand{\anonsubtext}{(No author info supplied here, for consistency with
TACL-submission anonymization requirements)}
\newcommand{\instr}
\fi

\iftaclpubformat % this "if" is set by the choice of options

\else

\fi

\title{MuISQA: Multi-Intent Retrieval-Augmented Generation for Scientific Question Answering}

\author{
  Zhiyuan Li\textsuperscript{1,2},
  Haisheng Yu\textsuperscript{1},
  Guangchuan Guo\textsuperscript{1},
  Nan Zhou\textsuperscript{1},
  Jiajun Zhang\textsuperscript{1,2,3,4}\Thanks{Corresponding author
  \\
  The proposed MuISQA dataset and related codes can be
found at https://github.com/Zhiyuan-Li-John/MuISQA
  } 
  \\
  \textsuperscript{1}Zhongke Zidong Taichu (Beijing), China
  \\
  \textsuperscript{2}Institute of Automation, Chinese Academy of Sciences, China
  \\
  \textsuperscript{3}School of Artificial Intelligence, University of Chinese Academy of Sciences, China
  \\
  \textsuperscript{4}Wuhan AI Research
  \\
  \texttt{\{lizhiyuan, yuhaisheng, guoguangchuan, zhounan\}@taichu.ai}
  \\
  \texttt{jjzhang@nlpr.ia.ac.cn}
}

\date{}

\begin{document}
\maketitle
\begin{abstract}
  Complex scientific questions often entail multiple intents, such as identifying gene mutations and linking them to related diseases. These tasks require evidence from diverse sources and multi-hop reasoning, while conventional retrieval-augmented generation (RAG) systems are usually single-intent oriented, leading to incomplete evidence coverage. To assess this limitation, we introduce the Multi-Intent Scientific Question Answering (MuISQA) benchmark, which is designed to evaluate RAG systems on heterogeneous evidence coverage across sub-questions. In addition, we propose an intent-aware retrieval framework that leverages large language models (LLMs) to hypothesize potential answers, decompose them into intent-specific queries, and retrieve supporting passages for each underlying intent. The retrieved fragments are then aggregated and re-ranked via Reciprocal Rank Fusion (RRF) to balance coverage across diverse intents while reducing redundancy. Experiments on both MuISQA benchmark and other general RAG datasets demonstrate that our method consistently outperforms conventional approaches, particularly in retrieval accuracy and evidence coverage.
\end{abstract}

\section{Introduction}
\begin{figure}[t]
  \centering
  \vspace{-0.0cm}   \includegraphics[width=1.0\linewidth]{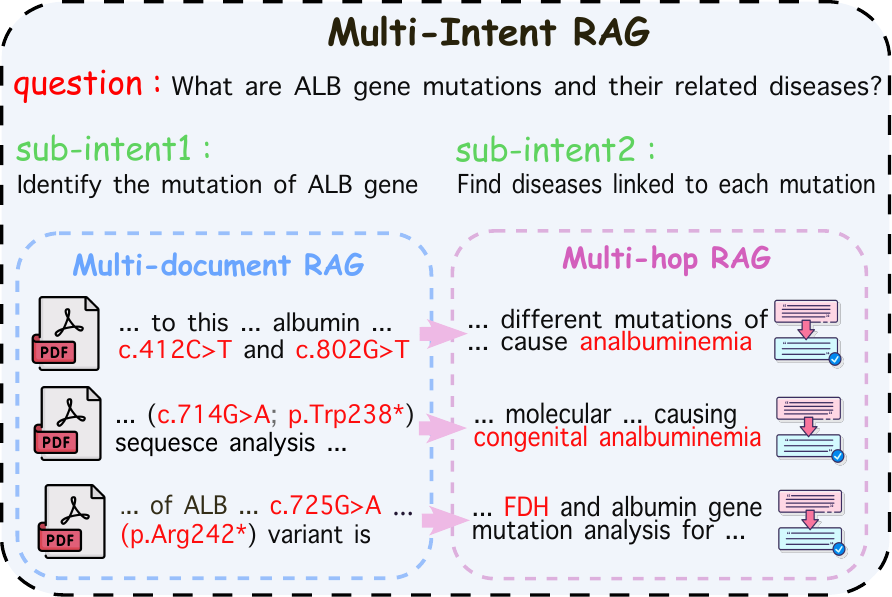}
   \vspace{-0.5cm}
   \caption{An example from our MuISQA benchmark, challenging RAG systems with multi-document retrieval and multi-hop reasoning.}
   \label{fig1}
   \vspace{-0.4cm}
\end{figure}
Retrieval-augmented generation (RAG) enhances Large Language Models (LLMs) with access to external knowledge sources, improving their factual reliability and cross-domain generalization~\cite{lee2019latent,karpukhin2020dense,mao2021generation}. At its core lies a retrieval module, typically implemented using dense retrieval techniques that exploit semantic embeddings to locate relevant evidence~\cite{gao2023precise,wang2023query2doc}. These approaches have achieved remarkable success on traditional question answering (QA) benchmarks~\cite{yang2018hotpotqa,kwiatkowski2019natural,ho2020constructing}, particularly with recent large-context embedding models such as Qwen3-Embedding-8B~\cite{zhang2025qwen3} or BGE-M3~\cite{multi2024m3}. 

Despite recent progress, most existing RAG systems~\cite{Zhao2024RetrievalAugmentedGF} and benchmarks~\cite{Singh2025AgenticRG} are built on a single-intent assumption, 
 where each question corresponds to one canonical answer, and retrieval performance is measured against that unique target. Widely used datasets such as TriviaQA~\cite{joshi2017triviaqa} and Natural Questions~\cite{kwiatkowski2019natural} further reinforce this paradigm by annotating a single gold span per query, with metrics like nDCG~\cite{jeunen2024normalised} and Recall@K~\cite{kynkaanniemi2019improved} designed accordingly. However, many scientific questions naturally entail multiple correlated intents. For example, a biomedical query about the human gene ALB may involve several mutations, each associated with different diseases. As illustrated in Figure~\ref{fig1}, answering such questions requires aggregating evidence across multiple documents and performing multi-hop reasoning for comprehensive coverage. Existing RAG systems~\cite{yu2024evaluation,fan2024survey} tend to focus on one dominant answer, repeatedly retrieving redundant evidence while overlooking complementary fragments that support alternative perspectives~\cite{xie2025rag}.

To systematically investigate this limitation, we introduce the \textbf{Mu}lti-\textbf{I}ntent \textbf{S}cientific \textbf{Q}uestion \textbf{A}nswering (MuISQA) benchmark, which is designed to evaluate how RAG systems handle questions with multiple correlated intents. MuISQA covers five scientific domains, including biology, chemistry, geography, medicine, and physics, with each question annotated for diverse sub-intents and their corresponding answers. Unlike prior benchmarks~\cite{kwiatkowski2019natural,ho2020constructing} that primarily emphasize precision or recall, MuISQA introduces evaluation metrics across three key dimensions: (i) \textbf{Query formulation}, measuring the ability to capture distinct intents; (ii) \textbf{Passage retrieval}, assessing coverage over different subtopics; and (iii) \textbf{Answer generation}, evaluating both accuracy and completeness of final responses. 

Building on these insights, we further develop an intent-aware retrieval framework that enhances both retrieval diversity and evidence coverage. It first leverages LLMs to hypothesize potential answers for the question and decomposes them into intent-specific queries. Unlike traditional query-rewriting methods~\cite{ma2023query,jagerman2023query} that generate semantically similar variants, our approach explicitly injects distinct hypothetical information into each query, broadening the search intent and improving evidence diversity. Compared with hypothetical document approaches such as HyDE~\cite{gao2023precise}, which rely on a single synthetic passage to guide retrieval, our method promotes retrieval diversification by decomposing multiple hypotheses into independent queries, each retrieving relevant passages for one intent via embedding similarity. The retrieved chunks are then aggregated and re-ranked using Reciprocal Rank Fusion (RRF)~\cite{cormack2009reciprocal}, which balances complementary evidence while reducing redundancy from single intent. 

We evaluate our framework on both the proposed MuISQA benchmark and several general RAG datasets, including TriviaQA~\cite{joshi2017triviaqa}, HotpotQA~\cite{yang2018hotpotqa}, NQ~\cite{kwiatkowski2019natural}, 2WikiMQA~\cite{ho2020constructing}, and MuSiQue~\cite{trivedi2022musique}, to assess its effectiveness and generalization. On MuISQA, our approach shows substantial gains in query efficiency and retrieval coverage over prior methods~\cite{ma2023query,gao2023precise,xie2025rag}, highlighting its ability to capture diverse intents and retrieve complementary evidence. Moreover, it surpasses recent state-of-the-art (SOTA) RAG systems~\cite{jimenez2024hipporag, shen2025gear, zhu2025mitigating} on general datasets, particularly on multi-hop reasoning tasks, demonstrating strong generalization. Interestingly, our analysis also uncovers that imperfect hypothetical generation, which is often regarded as LLM hallucination, can occasionally facilitate retrieval by guiding the search toward target passages, contrasting with the limitations previously reported for HyDE~\cite{gao2023precise}.

\section{Related Work}
\begin{figure*}[!th]
  \centering
  \vspace{-0.0cm}   \includegraphics[width=1.0\linewidth]{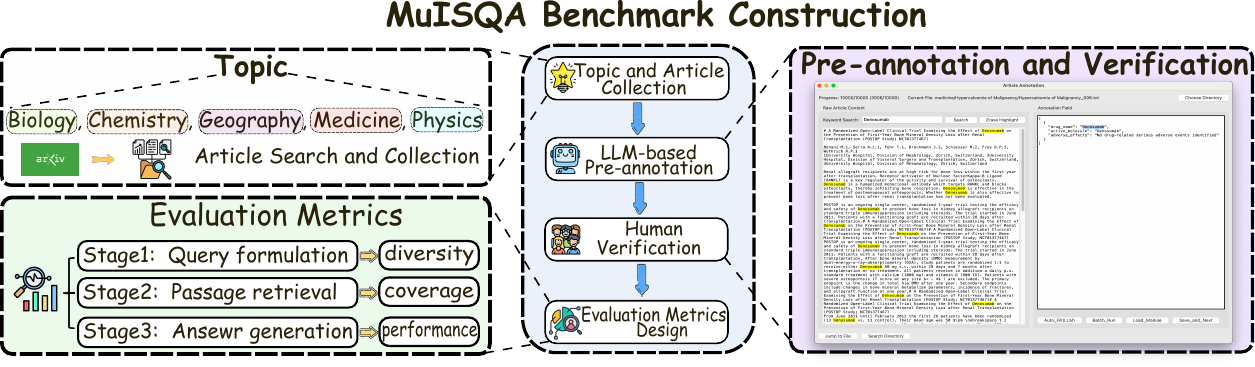}
   \vspace{-0.5cm}
   \caption{The structure of MuISQA benchmark construction. It follows synthesis in four core levels: (1) Topic and article collection, (2) LLM-based pre-annotation, (3) Human verification, and (4) Evaluation metrics design.}
   \label{fig2}
   \vspace{-0.2cm}
\end{figure*}

\subsection{RAG Benchmarks}
RAG benchmarks~\cite{yu2024evaluation,fan2024survey} have been central to evaluating systems across question answering~\cite{talmor2019commonsenseqa,chen2021finqa}, document retrieval~\cite{wang2024dapr,wasserman2025real}, and summarization~\cite{wang2020reviewrobot, petroni2021kilt}. Early datasets such as HotpotQA~\cite{yang2018hotpotqa} and TriviaQA~\cite{joshi2017triviaqa} largely follow a single-answer paradigm, emphasizing multi-hop reasoning~\cite{ho2020constructing,trivedi2022musique} and intensive knowledge~\cite{kwiatkowski2019natural} respectively. With advances in LLMs~\cite{liu2024deepseek,yang2025qwen3,openai2025gpt5} and embedding techniques~\cite{chen2024bge,zhang2025qwen3}, performance on these benchmarks has improved substantially. More recent works extend evaluation to multimodal~\cite{wasserman2025real,xia2024rule} and graph-structured retrieval tasks~\cite{xiao2025graphrag}, reflecting a shift toward more complex and realistic settings. Recently, a few studies have also begun exploring the multi-intent regime, where a question may contain multiple valid answers across distinct subtopics. For example, \citet{wang2025retrieval} investigates retrieval under conflicting or imbalanced evidence and finds that models tend to favor dominant answers, while \citet{xie2025rag} decomposes open-ended questions into sub-queries to assess evidence coverage. However, these efforts primarily focus on answer-side evaluation, offering limited insights into query diversification or retrieval coverage. In contrast, our benchmark directly targets the multi-intent setting and introduces metrics that disentangle performance across query formulation, evidence retrieval, and answer generation, providing a finer-grained diagnosis of system behaviour and clear guidance for improving RAG systems.

\subsection{Query Optimization in RAG}
Query optimization has been extensively explored as an approach to enhance retrieval accuracy in RAG systems~\cite{gao2023retrieval}. Early studies focused on query expansion and rewriting techniques, including classical relevance feedback methods such as Rocchio~\cite{rocchio1971relevance}, concept-based expansion~\cite{qiu1993concept}, and pseudo-relevance feedback~\cite{xu1996query}. Recent advances have shifted toward generative approaches, where LLMs synthesize richer or more diverse query representations. For instance, Query2Doc~\cite{wang2023query2doc} and HyDE~\cite{gao2023precise} generate pseudo-documents as expanded queries, DMQR-RAG produces rewrites with varying specificity to improve coverage~\cite{li2024DMQR}, and RaFe leverages reranker feedback to train more effective query rewrites~\cite{mao2024rafe}. Another recent work~\cite{xie2025rag} decomposes the question into core, background, and follow-up components to improve both retrieval coverage and answer generation. While these approaches share certain similarities and exhibit notable innovations in query optimization, they overlook the challenge of handling multiple underlying intents. In contrast, our framework hypothesizes diverse plausible answers, decomposes them into intent-specific queries, and aggregates retrieved chunks through RRF algorithm, directly addressing the multi-intent coverage problem rather than merely refining precision on a dominant intent.

\section{MuISQA Benchmark}
To construct MuISQA benchmark, we design a hybrid pipeline that integrates automated data acquisition with LLM-assisted annotation to efficiently build the dataset. As shown in Figure~\ref{fig2}, the construction process consists of four main stages: (1) \textbf{Topic and article collection}, where domain-relevant documents are gathered across multiple scientific disciplines; (2) \textbf{LLM-based pre-annotation}, which generates preliminary labels for each article through the LLM; (3) \textbf{Human verification and correction}, ensuring factual accuracy, coherence, and domain reliability in the final dataset; and (4) \textbf{Evaluation metrics design}, defining measures to assess system performance across query formulation, evidence retrieval, and answer generation. Representative examples from MuISQA are presented in Appendix~\ref{apx1}.

\subsection{Topic and Article Collection}
The MuISQA benchmark spans five scientific domains, including \textit{biology}, \textit{chemistry}, \textit{geography}, \textit{medicine}, and \textit{physics}. For each domain, we curate 100 multi-intent questions that admit multiple valid answers, resulting in 500 questions overall.  To construct the evidence corpus, representative keywords are extracted from each question and used to search on the arXiv website for relevant open-access articles. For every question, we retrieve 20 articles without manual filtering, collecting 10,000 documents in total. This setup mirrors realistic retrieval conditions, where relevant and irrelevant documents coexist, capturing the noise and diversity of real-world scenarios. 

\subsection{LLM-based Pre-annotation}
To enable LLM-assisted pre-annotation, we developed a dedicated annotation platform, as shown on the right side of Figure~\ref{fig2}. The system loads each question together with its retrieved documents, displaying the full article text on the left panel and the annotation interface on the right. We integrate the DeepSeek-V3~\cite{liu2024deepseek} into the tool and implement a one-click annotation function (\textit{Auto$\_$Fill (LLM)}), which automatically reads the document, combines it with the associated question, and generates a preliminary answer waiting for human review. Further implementation details are provided in Appendix~\ref{apx2}.

\subsection{Human Verification and Correction}
To support efficient human verification and correction, our platform integrates interactive features that streamline the validation of generated annotations. The left panel provides a keyword search function that highlights relevant terms within the document, helping annotators quickly locate evidence linked to the pre-annotated answers. When inconsistencies or missing details are found, annotators can directly edit the answers in the right interface. Once verification is complete, the results can be saved, and the annotator can move to the next document via the \textit{Save$\_$and$\_$Next} button. Further implementation details are provided in Appendix~\ref{apx3}.

\subsection{Evaluation Metrics Design}
Unlike previous RAG benchmarks that primarily focus on one canonical answer, each question in MuISQA involves multiple intents leading to multiple correct answers, with each answer potentially supported by several semantically similar passages drawn from different documents. This structure makes conventional metrics such as Exact Match (EM) or Recall inadequate for capturing system performance. To address this, we design tailored evaluation metrics specifically for multi-intent questions, disentangling model capability across three critical stages:
\begin{itemize}
\item Query formulation, evaluating the model’s ability to capture diverse underlying intents;
\item Passage retrieval, measuring the coverage of retrieved evidence across subtopics;
\item Answer generation, assessing both the factual accuracy and completeness of the final response.
\end{itemize}

For query formulation, rather than directly evaluating intent diversity, we introduce a metric called vector entropy to quantify the informational complexity of query representations. Given a query consisting of $m$ sentences with embedding vectors $\{v^{(1)}, \dots, v^{(m)}\}$, each vector is first normalized into a probability distribution over dimensions:
\begin{gather}
p^{(s)}_i = \frac{|v^{(s)}_i|}{\sum_j |v^{(s)}_j|},
\end{gather}
where $v^{(s)}_i$ represents the $i$-th component of the vector. $p^{(s)}_i$ represents the normalized probability of the $s$-th sentence on the $i$-th dimension. We then compute the vector entropy as:
\begin{gather}
    p_{\text{mix}} = \frac{1}{m}\sum_{s=1}^{m} p^{(s)},\\
    H_{\text{mix}} = -\sum p_{\text{mix}}\log p_{\text{mix}},
\end{gather}
where $p_{\text{mix}}$ represents the overall semantic distribution of the entire query, $H_{\text{mix}}$ indicates the diversity of embedding semantics.

\begin{figure*}[!th]
  \centering
  \vspace{-0.0cm}   
  \includegraphics[width=1.0\linewidth]{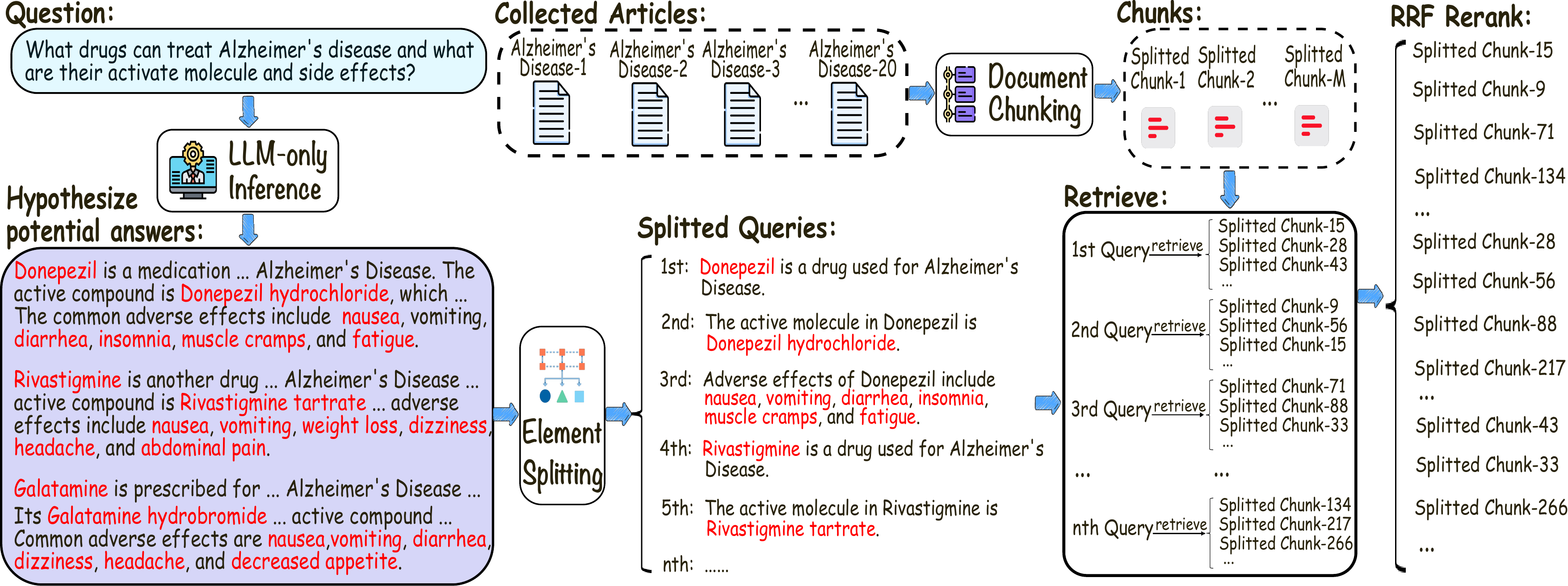}
   \vspace{-0.5cm}
   \caption{The overview of our proposed intent-aware retrieval framework. The LLM first generates hypothetical answers and decomposes them into diverse intent-specific queries. These queries are then used to retrieve relevant document chunks, which are re-ranked using the RRF algorithm to ensure comprehensive coverage of evidence.}
   \label{fig3}
   \vspace{-0.2cm}
\end{figure*}

For passage retrieval, unlike conventional RAG evaluations that simply compute recall based on the number of retrieved passages, multi-intent questions require assessing how well a system retrieves diverse and informative content. We therefore propose the Information Recall Rate (IRR), which measures the amount of distinct factual information recovered from retrieved passages. Specifically, IRR extracts all heterogeneous factual units within the retrieved set that align with the gold answers. Given a query 
with a set of retrieved passages $\mathcal{R} = \{r_1, \dots, r_n\}$ and the corresponding gold-standard answers represented as factual units $\mathcal{G} = \{g_1, \dots, g_k\}$, we first extract all factual units from the retrieved passages that align with the gold set, forming the subset $\mathcal{R^*} = \mathcal{R \cap G}$. The ratio between the number of correctly retrieved unique information units and the total number of gold information units is defined as:
\begin{gather}
\text{IRR} = \frac{|\mathcal{R}^*|}{|\mathcal{G}|},
\end{gather}
where $|\mathcal{R^*}|$ denotes the number of correctly retrieved unique factual units, and $|\mathcal{G}|$ represents the total number of gold-standard factual units. This metric reflects how effectively a retrieval method captures complementary evidence for multiple intents, rather than repeatedly retrieving redundant fragments for a single one.

For answer generation, we evaluate the quality of the final responses from two complementary perspectives: Answer Accuracy (AA) and Answer Coverage (AC). These metrics jointly capture how precise and comprehensive the generated answers are relative to the gold-standard set. Given a generated answer set $\mathcal{A}_{gen} = \{a^{(1)}, \cdots, a^{(N)}\}$ and the corresponding gold-standard answers $\mathcal{A}_{gold} = \{b^{(1)}, \cdots, b^{(K)}\}$. We first identify the subset of correctly generated answers that exactly or semantically match the gold annotations $\mathcal{A^*} = \mathcal{A}_{gen} \cap \mathcal{A}_{gold}$. We then define Answer Accuracy (AA) as the ratio of correct answers among all generated ones:
\begin{gather}
\text{AA} = \frac{|\mathcal{A}^*|}{|\mathcal{A}_{\text{gen}}|}.
\end{gather}
Similarly, we define Answer Coverage (AC) as the proportion of gold-standard answers that are successfully generated:
\begin{gather}
\text{AC} = \frac{|\mathcal{A}^*|}{|\mathcal{A}_{\text{gold}}|}.
\end{gather}
Together, AA and AC offer a balanced evaluation of precision and recall at the generation stage, complementing the earlier metrics on query formulation and passage retrieval to provide a comprehensive assessment of multi-intent RAG performance.

\section{Method}
In this section, we introduce our intent-aware retrieval framework. As shown in Figure~\ref{fig3}, the framework consists of two core components: Hypothetical Query Generation and RRF-based Reranking. The first module (Section~\ref{generation}) leverages LLMs to hypothesize potential answers for a multi-intent question and decompose them into diverse, intent-specific queries enriched with distinct hypothetical information. The second module (Section~\ref{reranking}) retrieves relevant document chunks for each generated query and re-ranks them using the RRF algorithm, promoting comprehensive and balanced evidence coverage across multiple intents.

\subsection{Hypothetical Query Generation}
\label{generation}
Query expansion~\cite{carpineto2012survey} has long been an effective strategy for enhancing retrieval performance. Recent research broadly falls into two paradigms. The first focuses on \textbf{query reformulation}, where complex questions are decomposed into finer-grained sub-questions or follow-up queries that capture distinct facets of the original intent~\cite{ma2023query,jagerman2023query}. The second, known as \textbf{hypothetical document generation}\cite{wang2023query2doc,gao2023precise}, instructs large language models (LLMs) to synthesize a pseudo-document that plausibly answers the query, using it as a semantic proxy during retrieval. Compared with direct rewriting, hypothetical document generation often yields superior performance in specialized domains, as the generated passages implicitly embed domain knowledge\cite{wang2023query2doc}, thereby enriching the query representation and improving embedding alignment with relevant evidence.

When dealing with multi-intent questions, generating a single hypothetical document often proves suboptimal. As illustrated on the left side of Figure~\ref{fig3}, a long synthetic passage is typically dominated by one prevailing intent (e.g., adverse drug effects) while underrepresenting others (e.g., therapeutic efficacy or active compounds). Using the entire passage embedding for retrieval risks emphasizing certain intents and suppressing complementary information, leading to incomplete evidence coverage. To overcome this limitation, we propose the Hypothetical Query Generation (HQG) method, which transforms an LLM-generated hypothetical answer into multiple intent-specific queries. Instead of encoding the entire passage as a single dense embedding, HQG explicitly decomposes it into a set of focused statements, each capturing a distinct sub-intent of the original question. Given a question $\mathcal{Q}$ with an underlying set of latent intents $\mathcal{I} = \{I_1, I_2, \cdots, I_L\}$, We first instruct the LLM to generate a hypothetical paragraph $\tilde{P} = \{\tilde{P}^{(1)}, \tilde{P}^{(2)}, \dots, \tilde{P}^{(M)}\}$, where each $\tilde{P}^{(m)}$ represents a distinct hypothetical answer instance that the model considers plausible for question $\mathcal{Q}$. For example, when asked "\textit{What drugs can treat Alzheimer's disease and what are their active molecule and side effects?}", each $\tilde{P}^{(m)}$ may describe one potential drug and its associated active molecule and side effects. Each hypothetical answer $\tilde{P}^{(m)}$ is further decomposed into a set of intent-specific factual statements:
\begin{gather}
    \tilde{P}^{(m)} \rightarrow \{ s^{(m)}_1, s^{(m)}_2, \dots, s^{(m)}_L \},
\end{gather}
where $s^{(m)}_\ell$ corresponds to the $\ell$-th intent for the $m$-th hypothetical answer. Following this, all decomposed statements form the complete hypothetical fact pool:
\begin{gather}
    \mathcal{S} = \bigcup_{m=1}^{M} \{ s^{(m)}_1, s^{(m)}_2, \dots, s^{(m)}_L \},
\end{gather}
where $\mathcal{S}$ represents the complete hypothetical query pool. $M$ represents the number of distinct hypothetical answer instances.

\subsection{RRF-based Reranking}
\label{reranking}
Once the complete hypothetical query pool is generated, each query is independently issued to the retrieval system, producing candidate passages relevant to different sub-intents. Each factual statement $s^{(m)}_\ell$ from the hypothetical query pool is first embedded into a dense representation:
\begin{gather}
    h^{(m)}_\ell = f_{\text{embed}}(s^{(m)}_\ell),
\end{gather}
where $f{\text{embed}}(\cdot)$ denotes the embedding function. Each embedding $h^{(m)}_\ell$ is then used to search the corpus via semantic similarity, returning a ranked list of top$\_K$ passages:
\begin{gather}
    \mathcal{R}^{(m)}_\ell = \text{Retrieve}(h^{(m)}_\ell) \\
    = \{ (d_i, \text{rank}^{(m,\ell)}_i, \text{score}^{(m,\ell)}_i) \}_{i=1}^{K},
\end{gather}
where $d_i$ represents the $i$-th retrieved chunks for the $s^{(m)}_\ell$ with rank $\text{rank}^{(m,\ell)}_i$ and $\text{score}^{(m,\ell)}_i$. This process yields multiple ranked lists corresponding to different hypothetical answers and intents.

\begin{table*}[t]
\centering
\scalebox{1.02}{
\footnotesize
\begin{NiceTabular}{lccccc}[
% code-before={
%     \rowcolor{rowgray}{6,11,16,21} % 先给4行整行上色
%     \cellcolor{white}{6-1}         % 再把这4行的第1列恢复为白色
%     \cellcolor{white}{11-1}
%     \cellcolor{white}{16-1}
%     \cellcolor{white}{21-1}
%   }, 
  cell-space-limits=1.5pt]
\specialrule{1.2pt}{0pt}{0.2pt}
LLMs                         & Methods                                                     & Average $H_{mix}$ & Average IRR & Average AA & Average AC \\ \hline
\multirow{5}{*}{Qwen2.5-72B} & Naive RAG~\cite{karpukhin2020dense}                                                   & 8.058                & 61.240      & 56.674     & 46.606     \\ 
                             & Query   Rewritting~\cite{ma2023query} & 8.089                & 63.068      & 54.842     & 47.196     \\  
                             & HyDE~\cite{gao2023precise}             & \textbf{8.170}                & 63.506      & 56.452     & 46.744     \\ 
                             & Core-SubQ~\cite{xie2025rag}                & 8.103                & 62.930      & 54.804     & 45.598     \\ 
                             & Our method                                                        & 8.136                & \textbf{67.502}      & \textbf{57.778}     & \textbf{49.078}     \\ \hline
\multirow{5}{*}{Qwen3-235B}  & Naive RAG~\cite{karpukhin2020dense}                                                    & 8.058                & 61.412      & 56.730     & 48.054     \\  
                             & Query   Rewritting~\cite{ma2023query} & 8.091                & 62.874      & 55.080     & 48.012     \\ 
                             & HyDE~\cite{gao2023precise}            & \textbf{8.183}                & 62.308      & 55.232     & 48.226     \\ 
                             & Core-SubQ~\cite{xie2025rag}                & 8.109                & 62.202      & 56.760     & 48.328     \\ 
                             & Our method                                                       & 8.155                & \textbf{66.364}      & \textbf{57.158}     & \textbf{50.546}     \\ \hline
\multirow{5}{*}{Deepseek-V3} & Naive RAG~\cite{karpukhin2020dense}                                                    & 8.058                & 61.342      & 53.958     & 45.772     \\ 
                             & Query   Rewritting~\cite{ma2023query} & 8.084                & 62.056      & 56.566     & 47.186     \\ 
                             & HyDE~\cite{gao2023precise}             & \textbf{8.181}                & 60.326      & 56.156     & 46.876     \\ 
                             & Core-SubQ~\cite{xie2025rag}                 & 8.096                & 62.094      & 56.334     & 46.526     \\ 
                             & Our method                                                        & 8.148                & \textbf{65.296}      & \textbf{59.498}     & \textbf{50.108}     \\ \hline
\multirow{5}{*}{Deepseek-R1} & Naive RAG~\cite{karpukhin2020dense}                                                    & 8.058                & 61.346      & 55.044     & 42.628     \\ 
                             & Query   Rewritting~\cite{ma2023query} & 8.083                & 61.074      & 54.682     & 42.616     \\ 
                             & HyDE~\cite{gao2023precise}            & \textbf{8.169}                & 63.524      & 54.812     & 43.382     \\ 
                             & Core-SubQ~\cite{xie2025rag}                & 8.105                & 62.592      & 55.402     & 44.688     \\ 
                             & Our method                                                       & 8.160                & \textbf{66.221}      & \textbf{57.902}     & \textbf{46.304}     \\ 
\specialrule{1.2pt}{0pt}{0.2pt}
\end{NiceTabular}}
\vspace{-0.0cm}
\caption{The performance of different RAG approaches on the MuISQA benchmark. “Average” denotes the mean score across five scientific domains.}
\vspace{-0.3cm}
\label{tab1}
\end{table*}

After generating multiple ranked lists corresponding to different hypothetical answers and intents, it becomes necessary to aggregate and re-rank them into a unified ranking. However, simply relying on raw similarity scores for this process can be problematic for two main reasons. First, the retrieved passages originate from diverse documents rather than a single source, and variations in writing style or contextual framing may cause substantial fluctuations in similarity scores, even among semantically equivalent evidence. Second, dense embedding models often exhibit bias toward certain fragment types, such as those containing high-frequency domain terms, assigning them disproportionately high similarity scores while undervaluing other relevant evidence. To address these issues, we adopt the RRF algorithm~\cite{cormack2009reciprocal}, which combines results based on their relative ranks rather than their raw similarity scores, thereby producing a more balanced and robust aggregation of heterogeneous evidence. Given multiple ranked lists $\{ \mathcal{R}^{(m)}_\ell \}$, RRF assigns to each retrieved passage $d_i$ an aggregate ranking score:
\begin{gather}
\text{RRF}(d_i) = \sum_{m=1}^{M} \sum_{\ell=1}^{L} \frac{1}{k + \text{rank}^{(m,\ell)}_i},
\end{gather}
where $\text{rank}^{(m,\ell)}_i$ denotes the rank position of $d_i$ in list $\mathcal{R}^{(m)}_\ell$, and $k$ is a smoothing constant that dampens the influence of low-ranked items. By focusing on ranking order rather than absolute similarity, RRF effectively balances evidence from diverse hypothetical queries and reduces the dominance of any single bias-prone query or document.

\section{Experiments}
\subsection{Experimental Setup}
We evaluate our approach on MuISQA benchmark, comparing it with recent Query expansion methods: Query Rewriting~\cite{ma2023query}, HyDE~\cite{gao2023precise}, and Core-SubQ~\cite{xie2025rag}, which is conceptually related to our framework. Each retrieval strategy is paired with LLMs of different capacities: Qwen2.5-72B-Instruct~\cite{bai2025qwen2}, Qwen3-235B-A22B~\cite{yang2025qwen3}, DeepSeek-V3-0324~\cite{liu2024deepseek}, and DeepSeek-R1-0528~\cite{guo2025deepseek}.
To examine generalization, we also test on general RAG benchmarks, including multi-hop QA datasets (HotpotQA~\cite{yang2018hotpotqa}, 2WikiMQA~\cite{ho2020constructing}, MuSiQue~\cite{trivedi2022musique}) and Intensive knowledge QA datasets (NQ~\cite{kwiatkowski2019natural}, TriviaQA~\cite{joshi2017triviaqa}). Across all experiments, the RRF smoothing constant is set to K=60 and the number of retrieved passages is fixed at k=10.

\subsection{Experimental Results on MuISQA}
Table~\ref{tab1} presents the overall performance of representative query-expansion RAG methods with different LLMs on the MuISQA benchmark, leading to the following observations: (1) In the query formulation stage, our framework generates richer and more diverse query representations than Query Rewriting and Core-SubQ, as indicated by higher $H_{mix}$ scores. Although its score is slightly lower than HyDE, this difference stems from our decomposition of a single long query into multiple shorter, intent-specific ones, resulting in finer but more fragmented representations. (2) In the retrieval stage, queries produced by our framework achieve more comprehensive coverage, yielding higher Information Recall Rate (IRR) scores. (3) In the answer generation stage, our framework consistently improves both Answer Accuracy and Answer Coverage, achieving average gains of 2.5$\%$ and 3.0$\%$, respectively, over the naive RAG baseline.

\subsection{Experimental Results on General RAG Benchmarks}
\begin{table*}[t]
\centering
\vspace{-0.2cm}
\scalebox{1.02}{
\footnotesize
\begin{NiceTabular}{lccccccccc}[cell-space-limits=1.7pt]
\specialrule{1.2pt}{0pt}{0.2pt}
                          & \multicolumn{3}{c}{HotpotQA}                      & \multicolumn{3}{c}{2WikiMQA}                      & \multicolumn{3}{c}{MuSiQue}                       \\ \cline{2-10} 
\multirow{-2}{*}{Methods} & EM    & F1    &  R@10 & EM    & F1    & R@10 & EM    & F1    &  R@10 \\ \hline
NaiveRAG~\cite{karpukhin2020dense}    & 40.10 & 57.60 & 89.00                             & 39.80 & 51.80 & 83.20                             & 8.20  & 18.20 & 47.00                             \\
IRCoT~\cite{trivedi2023interleaving}                     & 45.20 & 63.70 & 92.50                             & 32.40 & 43.60 & 86.60                             & 12.20 & 24.10 & 54.30                             \\
LongRAG~\cite{jiang2024longrag}                   & 46.50 & 60.29 & 93.40                             & 51.50 & 62.00 & 88.70                             & 29.50 & 40.89 & 57.90                             \\
HippoRAG~\cite{jimenez2024hipporag}         & 49.20 & 67.90 & 94.70                             & 45.60 & 59.00 & 90.60                             & 14.20 & 25.90 & 54.50                             \\
GEAR~\cite{shen2025gear}                      & 50.40 & 69.40 & 96.80                             & 47.40 & 62.30 & \textbf{95.30}                             & 19.00 & 35.60 & 67.60                             \\
ChainRAG~\cite{zhu2025mitigating}                  & 52.00 & 64.54 &  96.20                                 & 55.00 & 65.85 & 90.50                                 & 39.00 & 49.37 & 74.80                                \\
Our method                & \textbf{59.72} & \textbf{74.10} & \textbf{97.20}                             & \textbf{59.07} & \textbf{66.99} & 91.70                             & \textbf{44.03} & \textbf{57.06} & \textbf{89.84}                             \\ 
\specialrule{1.2pt}{0pt}{0.2pt}
\end{NiceTabular}}
\vspace{-0.0cm}
\caption{The performance of representative RAG approaches on HotpotQA, 2WikiMQA, and MuSiQue. “EM” denotes Exact Match, “F1” denotes F1 score, and “R@10” denotes Recall at top 10 retrieved passages.}
\vspace{-0.1cm}
\label{tab2}
\end{table*}

\begin{table}[t]
\centering
\scalebox{1.1}{
\footnotesize
\begin{NiceTabular}{lcccc}[cell-space-limits=1.3pt]
\specialrule{1.1pt}{0pt}{0.2pt}
\multirow{2}{*}{Methods} & \multicolumn{2}{c}{NQ} & \multicolumn{2}{c}{TriviaQA} \\ \cline{2-5} 
                         & EM         & F1        & EM            & F1           \\ \hline
Naive RAG                & 35.80      & 51.2      & 45.80         & 58.10        \\
COMPACT                  & 38.40      & 50.00     & 65.40         & 74.90        \\
SURE                     & 39.40      & 52.30     & 50.40         & 63.00        \\
Self-Selection      & 37.80      & 52.50     & 56.60         & 66.30        \\
Our Method               & 28.02      & 57.14     & 66.15         & 78.63        \\ 
\specialrule{1.1pt}{0pt}{0.2pt}
\end{NiceTabular}}
\caption{The performance of representative approaches on intensive knowledge benchmarks.}
\vspace{-0.3cm}
\label{tab3}
\end{table}

\paragraph{Multi-hop QA Datasets.} 
Although our framework is primarily designed for multi-intent question answering, it generalizes naturally to single-intent multi-hop reasoning tasks. For example, the HotpotQA question "Were Scott Derrickson and Ed Wood of the same nationality?" is decomposed by our framework into two retrieval queries: "Scott Derrickson is an American filmmaker" and "Ed Wood was an American filmmaker", which directly lead to the supporting passages describing each person’s nationality. We further evaluate on three widely used multi-hop QA benchmarks: HotpotQA~\cite{yang2018hotpotqa}, 2WikiMQA~\cite{ho2020constructing}, and MuSiQue~\cite{trivedi2022musique}. Table~\ref{tab2} compares our method with recent representative RAG approaches~\cite{karpukhin2020dense,trivedi2023interleaving,jiang2024longrag,jimenez2024hipporag,shen2025gear,zhu2025mitigating}. Overall, our framework consistently achieves the best results in both EM and F1 scores, with particularly notable gains on HotpotQA (+7.7$\%$ EM, +9.5$\%$ F1). Moreover, it delivers higher retrieval recall on most datasets (except 2WikiMQA), including a +15$\%$ improvement in R@10 on MuSiQue.

\begin{figure}[t]
  \centering
  \vspace{-0.0cm}   
  \includegraphics[width=0.98\linewidth]{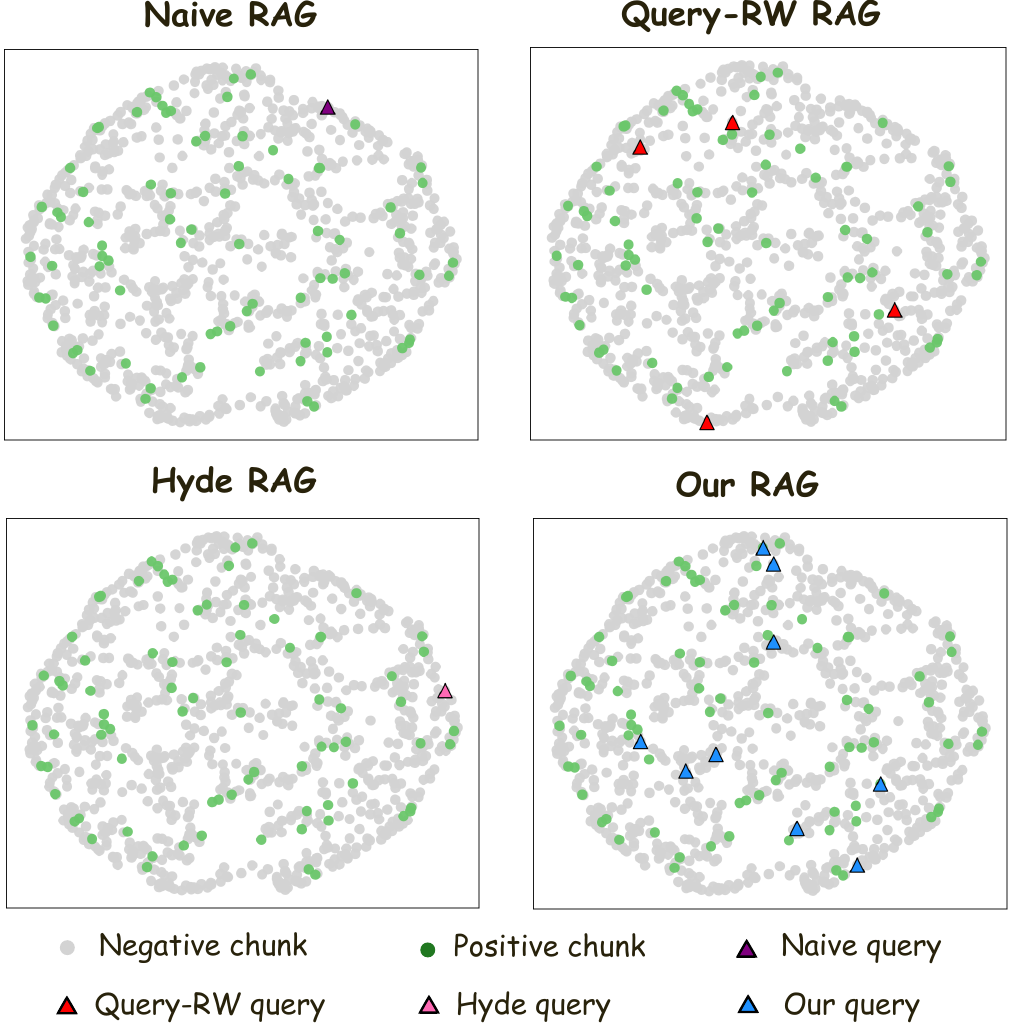}
   \vspace{-0.1cm}
   \caption{The UMAP visualization of an example from the MuISQA dataset. Best viewed by zooming in.}
   \label{fig4}
   \vspace{-0.3cm}
\end{figure} 

\paragraph{Intensive Knowledge QA Datasets.} 
We further evaluate our framework on intensive knowledge QA benchmarks, including NQ~\cite{kwiatkowski2019natural} and TriviaQA~\cite{joshi2017triviaqa}. Unlike multi-hop questions involving multiple entities or relations, intensive knowledge questions focus on factual details about a single subject. To adapt this, we modify the Hypothetical Query Generation process to split each answer into at least two independent queries.
For example, the NQ question "Who owned the Colts when they left Baltimore?" will generate two queries: "Robert Irsay was the owner of the Baltimore Colts." and "They left Baltimore in 1984." to improve retrieval accuracy. As shown in Table~\ref{tab3}, our method outperforms prior RAG approaches (COMPACT~\cite{yoon2024compact}, SURE~\cite{kim2024sure}, Self-Selection~\cite{weng2025optimizing}), achieving +4.6$\%$ and +3.7$\%$ F1 gains on NQ and TriviaQA, respectively.

\section{Discussion}
\subsection{Efficiency of the Intent-aware Retrieval}
We analyze why the proposed intent-aware retrieval framework achieves substantial gains in passage recall over prior RAG methods. Figure~\ref{fig4} shows a representative MuISQA case using UMAP, showing that our approach generates a larger and more semantically diverse set of queries that span broader regions of the embedding space. Compared with native RAG, HyDE and Query Rewriting, our queries are widely dispersed rather than clustered in narrow neighbourhoods. 

\begin{figure}[t]
  \centering
  \vspace{-0.0cm}   
  \includegraphics[width=1.0\linewidth]{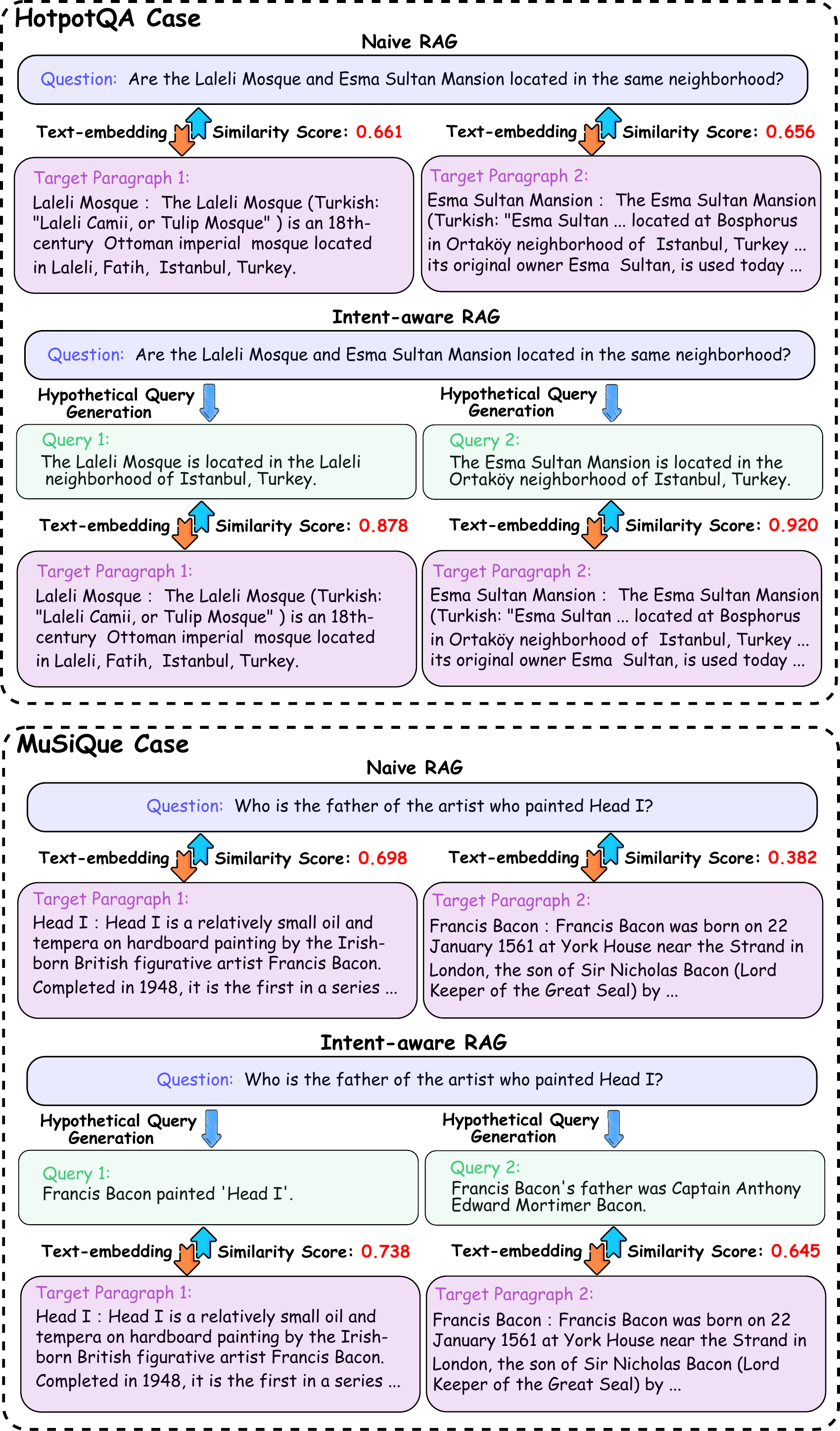}
   \vspace{-0.2cm}
   \caption{The case studies on HotpotQA and MusiQue benchmarks. Best viewed by zooming in.}
   \label{fig5}
   \vspace{-0.2cm}
\end{figure}

We further conduct case studies on multi-hop datasets to analyze retrieval behaviour in complex reasoning settings. As shown in Figure~\ref{fig5}, for the HotpotQA example, our framework leverages the LLM’s prior knowledge to infer the locations of the Laleli Mosque and Esma Sultan Mansion, effectively guiding retrieval to the correct passage. Conversely, hallucinated hypotheses can occasionally introduce factual errors, as in the MuSiQue case, where an incorrect statement about Francis Bacon’s father still increased similarity from 0.382 to 0.645 and led to successful retrieval, contrary to the limitations previously reported for HyDE~\cite{gao2023precise}.

\subsection{Impact of Hyperparameters}
\begin{figure}[t]
  \centering
  \vspace{-0.0cm}   
  \includegraphics[width=0.95\linewidth]{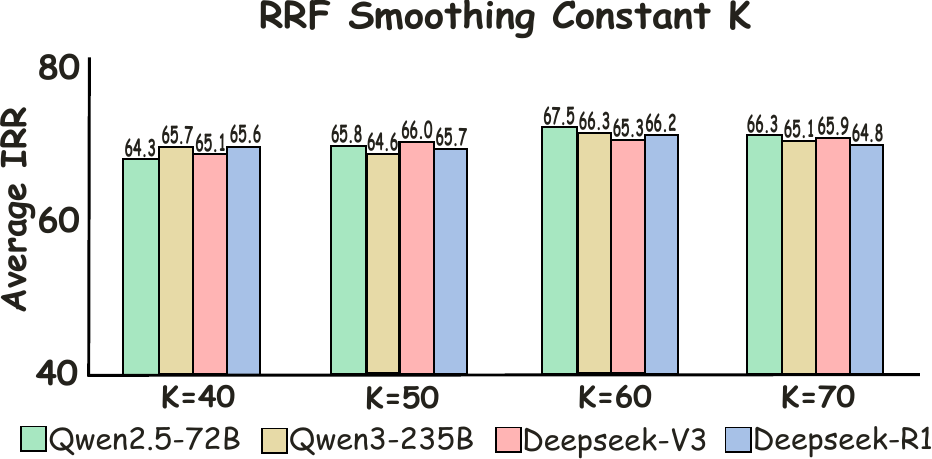}
   \vspace{-0.2cm}
   \caption{The performance comparison of different RRF Smoothing Constant K.}
   \label{fig6}
   \vspace{-0.2cm}
\end{figure}

\begin{figure}[t]
  \centering
  \vspace{-0.0cm}   
  \includegraphics[width=0.95\linewidth]{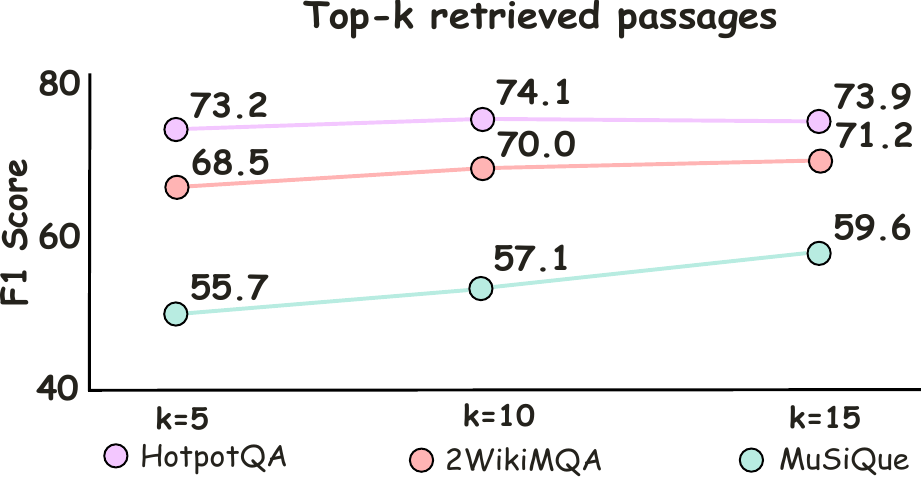}
   \vspace{-0.2cm}
   \caption{The performance comparison of different retrieval depth k.}
   \label{fig7}
   \vspace{-0.2cm}
\end{figure}

The performance of RAG systems is often sensitive to hyperparameter choices. We examine two key parameters in our framework: the smoothing constant K in the RRF algorithm and the number of retrieved passages k used for ranking. As shown in Figure~\ref{fig6}, varying K has only a minor effect on passage-level retrieval metrics, with optimal results achieved at K=60. We also analyze retrieval depth k (Figure~\ref{fig7}) and find that performance remains stable across a wide range of values. This robustness stems from our retrieval design, which consistently boosts the similarity scores of target passages and promotes their higher ranking during aggregation.

\begin{table}[!t]
\begin{NiceTabular}{lcccc}[cell-space-limits=1.3pt]
\specialrule{1.1pt}{0pt}{0.2pt}
Model       & $H_{mix}$ & IRR & AA & AC \\
\hline
Qwen2.5-3B  & 8.08          & 62.59    & 51.36   & 41.47   \\
Qwen2.5-7B  & 8.08          & 62.75    & 52.02   & 42.34   \\
Qwen2.5-72B & 8.13          & 67.50   & 57.78  & 49.08  \\
Qwen3-235B  & 8.15          & 66.36   & 57.16  & 50.55  \\ 
\specialrule{1.1pt}{0pt}{0.2pt}
\end{NiceTabular}
\caption{The performance of different generative models with our RAG approach on MuISQA benchmark.}
\label{tab3}
\vspace{-0.3cm}
\end{table}

\begin{figure*}[!th]
  \centering
  \vspace{-0.0cm}   \includegraphics[width=0.82\linewidth]{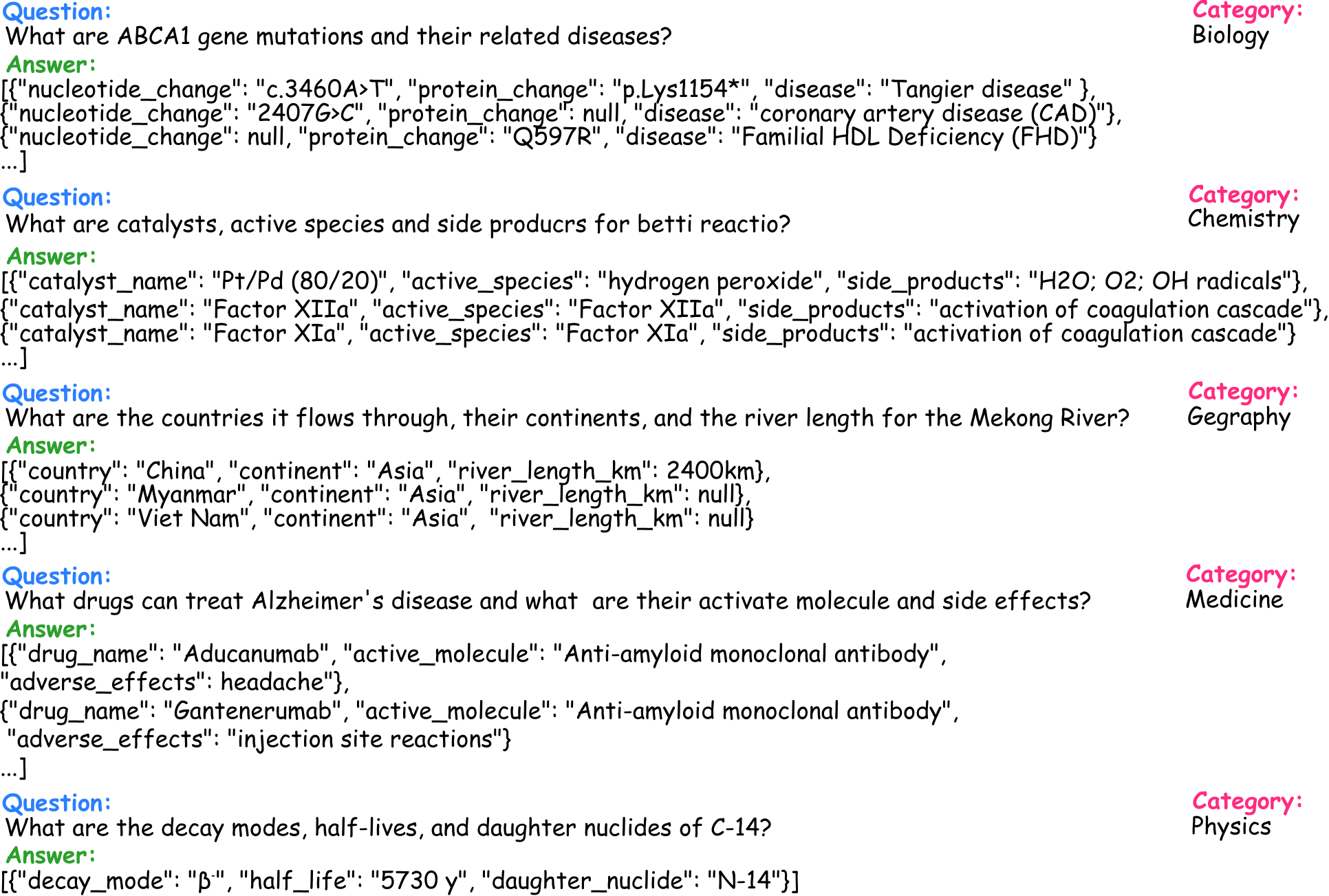}
   \vspace{-0.0cm}
   \caption{Examples from the MuISQA benchmark, each representing a multi-intent question with multiple valid answers across distinct subtopics. Best viewed by zooming in.}
   \label{fig8}
   \vspace{-0.0cm}
\end{figure*}

\subsection{Choice of Generative Models}
We examine how the capacity of the generative model influences the overall performance of our framework. Specifically, four Qwen variants: Qwen2.5-3B, Qwen2.5-7B, Qwen2.5-72B, and Qwen3-235B are used as LLMs for hypothetical answer generation. Table~\ref{tab3} shows that smaller models often produce less informative and domain-aware hypotheses, yielding weaker queries and reduced retrieval and answer quality. In contrast, larger models exhibit stronger factual grounding and richer knowledge priors, enabling more diverse and effective hypothetical statements that better guide retrieval.

\section{Conclusion}
In this work, We addressed the challenge of answering complex scientific questions that entail multiple correlated intents and require evidence from diverse sources and multi-hop reasoning. To systematically study this problem, we introduced MuISQA, a benchmark designed to evaluate RAG systems under multi-intent conditions. Building on this benchmark, we proposed an intent-aware retrieval framework that leverages LLMs to hypothesize potential answers, generate intent-specific queries, and fuse retrieved evidence via RRF. Experimental results demonstrate that our framework substantially enhances evidence diversity, retrieval coverage, and answer completeness compared with existing RAG baselines on both the MuISQA benchmark and the general dataset.

\section{Appendix}
\subsection{MuISQA Examples}
\label{apx1}
To illustrate the structure and diversity of MuISQA, Figure~\ref{fig7} presents representative examples from five scientific domains: \textit{biology}, \textit{chemistry}, \textit{geography}, \textit{medicine}, and \textit{physics}. Each example represents a multi-intent question with multiple valid answers covering distinct subtopics. 

\subsection{LLM-based Pre-annotation}
\label{apx2}
We developed a custom annotation interface and adopted DeepSeek-V3 as the sole model for pre-annotation, balancing efficiency and accuracy. Compared with the slower "think" models DeepSeek-R1 and Qwen3-235B, Deepseek-V3 offered a practical trade-off for large-scale processing. For long documents, articles were segmented into ~100K-token chunks to fit the 128K-token limit, each combined with its question and prompt. The generated answers were directly displayed for streamlined human verification.

\subsection{Human Verification}
\label{apx3}
Five annotators participated in dataset verification, each reviewing 200 questions: 100 unique and 100 overlapping for cross-validation. Their annotations were later aggregated: single reviews were adopted directly, while overlapping cases were cross-checked and merged into a unified version. As shown in Figure~\ref{fig2}, the platform also provides a \textit{Load$\_$Module} function for cases where LLM-generated annotations are inconsistent with the source text, enabling annotators to reload an empty template and conduct manual annotation.

\bibliography{tacl2021}
\bibliographystyle{acl_natbib}

\iftaclpubformat

\onecolumn

\appendix

\fi

\end{document}